\documentclass[runningheads]{llncs}
\usepackage[T1]{fontenc}
\usepackage{orcidlink}
\usepackage{graphicx,verbatim}
\usepackage{amsfonts}
\usepackage{amsmath} 
\usepackage{marvosym}
\usepackage[normalem]{ulem}
\useunder{\uline}{\ul}{}
\usepackage{url}
\usepackage{color}

\usepackage{booktabs} 
\usepackage{multirow} 
\usepackage{amssymb}
\begin{document}
\title{Gaussian Primitive Optimized Deformable Retinal Image Registration}

\author{
Xin Tian \inst{1} \and  
Jiazheng Wang \inst{3} \and 
Yuxi Zhang \inst{3} \and 
Xiang Chen \inst{3} \and 
Renjiu Hu \inst{1}  \and 
Gaolei Li \inst{4} \and 
Min Liu  \inst{3}  \and  
Hang Zhang \inst{2} \textsuperscript{(\Letter)} 
}
\authorrunning{X. Tian \emph{et al}.}

\institute{
University of Oxford, Oxford, UK  \and 
Cornell University, Ithaca, USA \\ \email{hz459@cornell.edu} \and
Hunan University, Changsha, China \and
Shanghai Jiao Tong University, Shanghai, China
}
    
\maketitle              
\begin{abstract}
Deformable retinal image registration is notoriously difficult due to large homogeneous regions and sparse but critical vascular features, which cause limited gradient signals in standard learning-based frameworks. In this paper, we introduce Gaussian Primitive Optimization (GPO), a novel iterative framework that performs structured message passing to overcome these challenges. After an initial coarse alignment, we extract keypoints at salient anatomical structures (e.g., major vessels) to serve as a minimal set of descriptor-based control nodes (DCN). Each node is modelled as a Gaussian primitive with trainable position, displacement, and radius, thus adapting its spatial influence to local deformation scales. A K-Nearest Neighbors (KNN) Gaussian interpolation then blends and propagates displacement signals from these information-rich nodes to construct a globally coherent displacement field; focusing interpolation on the top (K) neighbors reduces computational overhead while preserving local detail. By strategically anchoring nodes in high-gradient regions, GPO ensures robust gradient flow, mitigating vanishing gradient signal in textureless areas. The framework is optimized end-to-end via a multi-term loss that enforces both keypoint consistency and intensity alignment. Experiments on the FIRE dataset show that GPO reduces the target registration error from 6.2\,px to ~2.4\,px and increases the AUC at 25\,px from 0.770 to 0.938, substantially outperforming existing methods. The source code can be accessed via \url{https://github.com/xintian-99/GPOreg}.

\keywords{Retinal Vessel Alignment \and Deformable Image Registration \and Gaussian Primitive Parametrization \and Sparse Feature Propagation}

\end{abstract}
\section{Introduction}
\label{sec:intro}
\begin{figure}[t]
    \centering    \includegraphics[width=\linewidth]{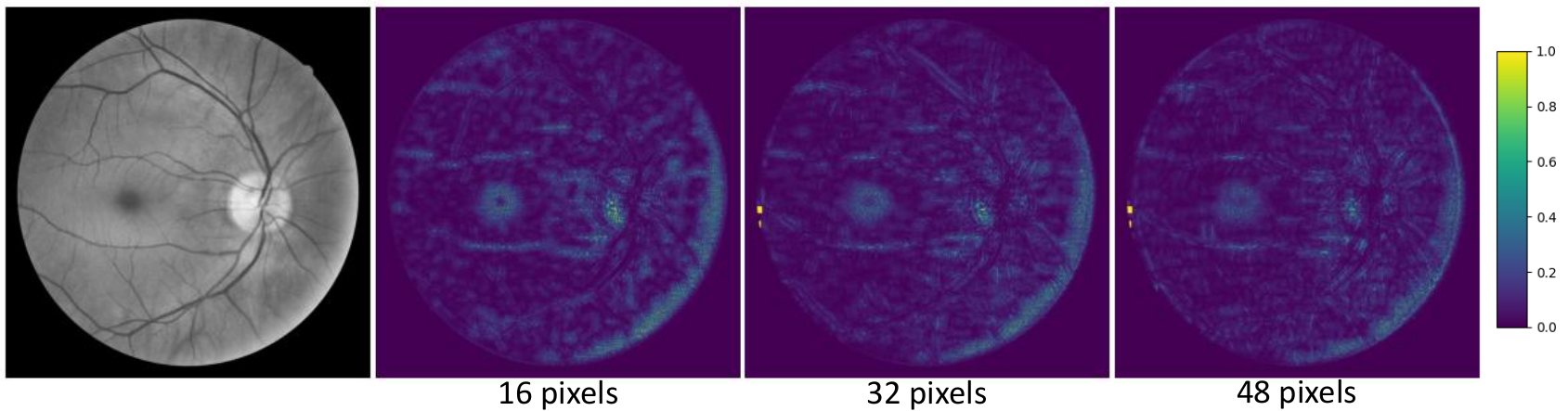}
    \caption{Visualization of gradient backflow in a retinal image under normalized cross-correlation (NCC). The original image (left) is preprocessed and normalized to [0,1]. Heatmaps (right) show the absolute NCC gradients for x-axis shifts of 16, 32, and 48 pixels. High responses indicate effective gradient propagation; low responses correspond to homogeneous or vessel-sparse regions.}
    \label{fig:sparsefeature}
\end{figure}

Retinal image registration is central to many clinical and research applications, including longitudinal monitoring of diseases such as diabetic retinopathy or age-related macular degeneration, as well as multi-modal fusion for enhanced diagnostic accuracy~\cite{noyel2017superimposition,tian2024tagat}. However, deformable retinal registration remains difficult due to the dominance of large homogeneous (textureless) regions and the sparse distribution of salient vasculature (less than 15\% of the area)~\cite{hu2021automatic}, which provides limited gradient signals and hampers accurate correspondence. As shown in Fig.~\ref{fig:sparsefeature}, the scarcity of vessel edges leads to weak gradients, while large flat regions offer little to no signal, presenting a challenge for both classical and learning-based registration methods.

Traditional registration methods iteratively optimize a similarity metric (e.g., normalized cross-correlation (NCC), or mutual information (MI)) using gradient-based approaches~\cite{avants2009advanced,klein2009elastix}. However, they are prone to local minima, and even advanced discrete optimization techniques~\cite{heinrich2013mrf,tian2022optimal,tian2019multimodal} often fail when images contain the extensive homogeneous regions and sparse vascular structures characteristic of the retina.

Modern deep learning-based approaches have also struggled to overcome this issue, broadly falling into three paradigms. (i) \textit{Regression-based} methods directly predict transformation parameters in a single forward pass (e.g., affine or flow predictors~\cite{chen2024spatially,de2019deep,mok2022affine,zhang2024slicer,zhao2019unsupervised}) for a coarse global alignment, but fail to model fine, local deformations and are susceptible to vanishing gradients in textureless regions. (ii) \textit{Descriptor-based} methods~\cite{detone2018superpoint,edstedt2024roma,liu2023geoformer,liu2022superretina,potje2024xfeat,revaud2019r2d2,truong2019glampoints} detect and match salient keypoints (usually on vessel junctions or other distinctive structures) to guide the transformation, but typically compute a single global transformation (e.g., homography) lacking an explicit data-fidelity term to refine local misalignments. More advanced (iii) \textit{learning and iterative optimization based} frameworks ~\cite{heinrich2019pdd,heinrich2022voxelmorph++,mok2021large,qiu2022GraDIRN,sivaraman2025retinaregnet,zheng2022recursive} integrate neural networks into multi-scale or iterative optimization pipelines. However, their reliance on dense image similarity or a simple smoothness regularizer lets loss function become dominated by easily aligned homogeneous regions, causing the crucial gradient signal from fine structures "diluted" or averaged out. Thus, thin vessels and other subtle anatomical features remain insufficiently registered due to restricted gradient backpropagation. 

In summary, classical optimization methods can become trapped in local minima, while descriptor-based solutions often compute a single global transformation, lacking the flexibility for local refinement. Furthermore, modern learning-based approaches suffer from gradient signal dilution, as the ambiguous displacement estimation in vast, textureless regions causes their loss to overwhelm the critical alignment signals from the sparse vasculature. A robust solution must therefore facilitate message passing to propagate displacement information from high-confidence vascular structures to these ambiguous regions in the network design~\cite{zhang2024memwarp,zhang2025smoothproper,zhang2023spatially}. To address these challenges, we propose Gaussian Primitive Optimization (GPO), a deformable registration framework designed specifically for sparse-feature medical images. Our key contributions are:
\begin{itemize}
\item We address gradient signal dilution by anchoring a minimal set of descriptor-based control nodes at salient keypoints, such as major vessels, which preserves the crucial displacement information provided by vascular structures.
\item Our KNN-based Gaussian interpolation serves as a structured message passing mechanism, propagating displacement signals from the control nodes into feature-sparse regions and blending local transformations into a globally coherent and locally precise alignment.
 \item On the FIRE dataset, GPO lowers the target registration error from 6.20\,px to 2.35\,px and increases the AUC@25\,px from 0.770 to 0.938, demonstrating both higher accuracy and fewer outlier errors compared to previous methods.
\end{itemize}

\section{Methodology}
\label{sec:method}
We propose a Gaussian Primitive Optimized (GPO) framework for retinal image registration, organized into four main stages. First, we perform coarse alignment using a descriptor-based network, which also yields matched keypoints to serve as control nodes. Next, each node is initialized as a Gaussian primitive with learnable position, displacement, and an adaptive radius. Then, we compute the deformation field via a structured message passing process, where a KNN Gaussian interpolation blends local transformations to accommodate complex retinal deformations. Finally, all parameters are iteratively optimized under a multi-term loss that enforces both intensity alignment and global consistency.

Formally, given a fixed image \(I_f\colon\Omega \to \mathbb{R}\) and a moving image \(I_m\colon \Omega \to \mathbb{R}\), we seek a transformation \(\mathcal{T}\colon \mathbb{R}^2 \to \mathbb{R}^2\) satisfying
\begin{equation}
I_f(\mathbf{x}) \approx I_m\bigl(\mathbf{x} + \mathbf{u}(\mathbf{x})\bigr),
\end{equation}
where \(\mathbf{u}(\mathbf{x})\) is the displacement field our GPO optimizes.
\begin{figure}[t]
    \centering
    \includegraphics[width=\textwidth]{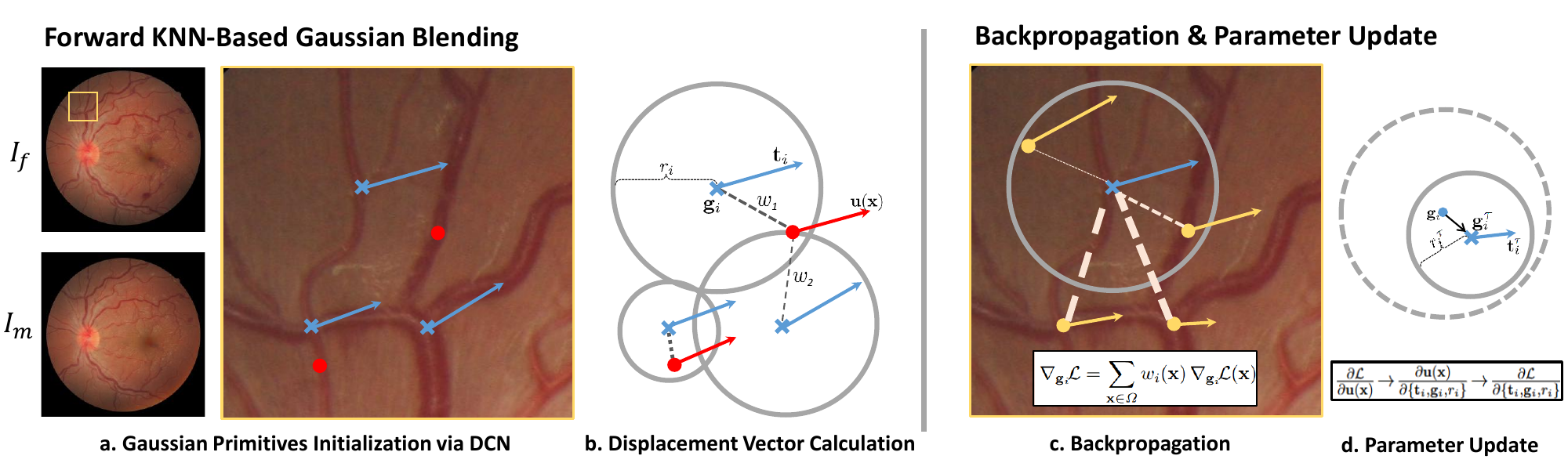}
    \caption{Overview of GPO: control node initialization, KNN-based Gaussian blending, and iterative parameter updates.}
    \label{fig:method}
\end{figure}

\subsection{Coarse Alignment \& Control Node Initialization}
We first obtain a coarse alignment of \( I_m \) to \( I_f \) by estimating a global transform:

\begin{equation}
I_m^{(\text{coarse})}(x) = I_m(A x + b),
\end{equation}

\noindent where \(A\) and \(b\) represent an affine, homography, or other global parameters learned by a descriptor-based network (e.g., GeoFormer~\cite{liu2023geoformer}). Concurrently, the network provides \(N\) matched keypoints \(\{(g_i^f, g_i^m)\}_{i=1}^N\) in \(I_f\) and \(I_m^{(\text{coarse})}\), forming \emph{descriptor-based control nodes (DCN)} (Fig.~\ref{fig:method}), which act as the primary sources for propagating displacement signals. If descriptors are unavailable or sparse, we sample \emph{grid-based control nodes (GCN)} on an \(n\times n\) lattice; each lattice point \(\{g_i\}_{i=1}^{n^2}\) in \(I_f\) is mapped to the same coordinate in \(I_m^{(\text{coarse})}\).

\subsection{KNN-Based Gaussian Blending for Deformation Estimation}
\subsubsection{Gaussian Primitive Initialisation}
After coarse alignment, each matched pair \( (g_i^f, g_i^m) \) is used to initialise a Gaussian primitive centered at \(\mathbf{g}_i \equiv g_i^f\). Every node~\(i\) possesses three sets of learnable parameters:

i. \textbf{Position} \(\mathbf{g}_i \in \mathbb{R}^2\): Refined during training to allow local anchors to shift toward anatomically salient regions.
    
ii. \textbf{Displacement Vector} \(\mathbf{t}_i \in \mathbb{R}^2\): Encodes the local translation of node \(\mathbf{g}_i\). Initialized to \(\mathbf{t}_i^{(0)} = g_i^m - g_i^f\) for DCN or \(\mathbf{0}\) for GCN.
    
iii. \textbf{Radius} \(r_i \in \mathbb{R}^+\): Adjusts each node’s spatial influence and parametrize \(r_i\) by a learnable scalar \(\beta_i\), via mapping
    \(
    r_i = r_{\min} + (r_{\max} - r_{\min}) \,\sigma(\beta_i) + 0.1,
    \)
    where \(\sigma(\cdot)\) is the sigmoid function. The 0.1 offset ensures \(r_i\) never becomes zero, preventing vanishing gradients.

\subsubsection{KNN-Based Gaussian Blending}

To propagate displacement signals and construct a smoothly varying displacement field $\mathbf{u}(\mathbf{x})$ from the sparse set of control nodes (Fig~\ref{fig:method} b), we employ a KNN-based Gaussian weighting scheme that functions as a message passing mechanism. Specifically, the displacement field is computed as a weighted sum over the \( K \)-nearest control nodes. We define the displacement at $\mathbf{x}$ as:
\begin{equation}
\label{eq:displacement-field}
\mathbf{u}(\mathbf{x}) = \sum_{i=1}^{K} w_i(\mathbf{x}) \mathbf{t}_i, 
\quad
\text{where} 
\quad
w_i(\mathbf{x}) = \frac{\exp \left(-\frac{\|\mathbf{x} - \mathbf{g}_i\|^2}{2 r_i^2} \right)}{\sum_{j=1}^{K} \exp \left(-\frac{\|\mathbf{x} - \mathbf{g}_j\|^2}{2 r_j^2} \right)},
\end{equation}
\noindent and $\mathbf{t}_i$ is the displacement vector associated with the $i$-th control node. The $w_i(\mathbf{x})$ is a Gaussian kernel weight that decays exponentially, thus giving greater influence to control nodes closer to $\mathbf{x}$. The denominator ensures $\sum_{i=1}^{K} w_i(\mathbf{x})=1$. Consequently, each pixel is influenced by its \(K\) nearest nodes, and each node, in turn, receives gradient signals from those pixels during backpropagation (see Sec.\,\ref{sec:backprop}). In descriptor-based control nodes (DCN), these nodes are placed on anatomically distinctive features (e.g., major vessels), preserving high-gradient signals to pixel and during optimization vice versa and mitigating vanishing gradients in homogeneous regions. Notably, by restricting on the top $K$ nearest nodes instead of all nodes, we reduce computational overhead while retaining accurate local detail. 

\subsection{Neural Iterative Optimization}

\subsubsection{Forward Pass \& Loss Function}
To find an optimal set of node parameters \(\{\mathbf{g}_i, \mathbf{t}_i, r_i\}_{i=1}^N\), we adopt an iterative, gradient-based framework indexed by \(\tau=1,\dots,\tau_{\max}\). At iteration \(\tau\), we compute the displacement field \(\mathbf{u}_{\tau}(\mathbf{x})\) from Eq.\,\eqref{eq:displacement-field} and warp the coarse-aligned moving image via bilinear interpolation for pixel resampling.
\begin{equation}
\label{eq:warp}
    I_{w,\tau}(\mathbf{x})
    \;=\;
    I_m^{(\text{coarse})}\!\bigl(\mathbf{x} + \mathbf{u}_{\tau}(\mathbf{x})\bigr),
\end{equation}

We learn the parameters $\{\mathbf{g}_i, \mathbf{t}_i, r_i\}_{i=1}^N$ and optimise the displacement field by minimising a two-term loss function:

\begin{equation}
\label{eq:loss}
\mathcal{L}_{\tau}
\;=\;
\alpha_1\,\mathcal{L}_{\mathrm{gcc}} \;+\; \alpha_2\,\mathcal{L}_{\mathrm{ncc}},
\end{equation}

\noindent where $\mathcal{L}_{\mathrm{gcc}}$ is global cross-correlation loss with matched control nodes, $\mathcal{L}_{\mathrm{ncc}}$ aligns overall intensity patterns in $I_f$ and $I_{w}^{(\tau)}$.

\subsubsection{Backpropagation \& Parameter Update}
\label{sec:backprop}
As \(\mathbf{u}(\mathbf{x})\) is a weighted sum over \(K\) nearest nodes (Eq.~\ref{eq:displacement-field}), each node \(\mathbf{g}_i\) accumulates gradient signals from multiple pixels during backpropagation via 
\(\tfrac{\partial \mathcal{L}}{\partial \mathbf{u}(\mathbf{x})}
\!\to\!
\tfrac{\partial \mathbf{u}(\mathbf{x})}{\partial \{\mathbf{t}_i, \mathbf{g}_i, r_i\}}
\!\to\!
\tfrac{\partial \mathcal{L}}{\partial \{\mathbf{t}_i, \mathbf{g}_i, r_i\}}\) to update \(\{\mathbf{g}_i, \mathbf{t}_i, r_i\}\) (Fig~\ref{fig:method} c \& d).
If \(\nabla_{\mathbf{g}_i}\!\mathcal{L}(\mathbf{x})\) denotes the pixel-level gradient at \(\mathbf{x}\) for node~\(i\), then its total gradient:
\begin{equation}
\nabla_{\mathbf{g}_i}\!\mathcal{L}
=
\sum_{\mathbf{x} \in \Omega}
w_i(\mathbf{x})\,\nabla_{\mathbf{g}_i}\!\mathcal{L}(\mathbf{x}),
\end{equation}
\noindent where \(w_i(\mathbf{x})\) is the Gaussian blending weight. Thus, even if vessels occupy only a small portion of the image, a subset of pixels near each node typically lies on or around high-gradient vessel edges, ensuring that every node still receives nonzero gradient signal for optimization and enabling robust convergence despite sparse vascular structures.

The \(\{\mathbf{g}_i, \mathbf{t}_i, r_i\}\) is then updated by subtracting their respective gradient terms scaled by distinct learning rates \(\eta_g, \eta_t, \eta_r\) at each iteration \(\tau\).
After \(\tau_{\max}\) iterations, we obtain the final displacement field \(\mathbf{u}_{\text{final}}\) for the coarse-aligned image to produce the fully registered result \( I_{w,\text{final}}(\mathbf{x})  =        I_m^{(\text{coarse})}\!\bigl(\mathbf{x} + \mathbf{u}_{\text{final}}(\mathbf{x})\bigr).
     \)

\section{Experiments, Results, and Discussion}
\label{sec:experiment}
\subsection{Experimental Setup and Baselines}
\label{sec:data}
\label{sec:setup}
\textbf{Dataset:}
We evaluated our approach on the FIRE dataset~\cite{hernandez2017fire}, which contains 134 retinal image pairs with 10 expert-annotated landmarks each for evaluation at a resolution of \(2912\times2912\) pixels. The dataset was previously collected and published in accordance with institutional ethical standards and the Declaration of Helsinki; no new human data collection was conducted for this study. The dataset has three subgroups: 71 pairs with minimal distortion (Category S), 4 pairs with anatomical changes (A), and 49 pairs with perspective distortion (P). Each image pair includes 10 expert-annotated landmark points for evaluation. For all experiments, we resized images to 1024×1024 and applied Gaussian blur for anti-aliasing.

\textbf{Baseline Methods:} We benchmark GPO against two categories of methods, with full results in Table~\ref{tab:comparison}: (1) descriptor-based methods for global transformation, including SuperPoint~\cite{detone2018superpoint}, R2D2~\cite{revaud2019r2d2}, RoMa~\cite{edstedt2024roma}, GeoFormer~\cite{liu2023geoformer}, and others; and (2) learning-based deformable registration frameworks, such as GraDIRN~\cite{qiu2022GraDIRN}, PDD-Net~\cite{heinrich2019pdd}, VoxelMorph++~\cite{heinrich2022voxelmorph++}, and RetinaRegNet~\cite{sivaraman2025retinaregnet}. For trainable methods, we used a 7:1:2 stratified train-val-test split. In qualitative comparisons, we highlight GeoFormer against our GPO-GCN and GPO-DCN variants.

\textbf{Evaluation Metrics:} We used two common metrics to evaluate model performance: Target Registration Error (TRE) and Area Under the Curve (AUC) for TRE thresholds. TRE measures the $L_2$ distance between corresponding points in the fixed and warped moving images annotated by clinical experts, with lower values indicating better alignment. AUC quantifies the percentage of TRE values below error thresholds (15, 25, and 50 pixels), normalizing accumulated rates to provide a comprehensive measure of registration success.

\textbf{Implementation Details:} The experiments were conducted in PyTorch and optimized on an NVIDIA A100 GPU using the Adam optimizer. The node position updates used an initial learning rate of \(\eta_g=1.0\), while both radii and displacement vectors used \(\eta_r=\eta_t=0.01\), enabling rapid coarse adjustments while preserving local fidelity. We fixed the KNN interpolation parameter to \(K=10\) for all experiments. For GPO-DCN, we sampled \(N=1000\) keypoints and optimized for 100 iterations; for GPO-GCN, we placed a \(20\times20\) grid of nodes and converged within 200 iterations. The loss weights in Eq.~\eqref{eq:loss} were set to \(\alpha_{\mathrm{gcc}}=0.4\) and \(\alpha_{\mathrm{ncc}}=1.0\), providing balanced guidance from both keypoint consistency and intensity similarity.

\subsection{Results and Analysis}

\subsubsection{Quantitative Analysis}
Table~\ref{tab:comparison} shows that both GPO variants substantially outperform selected descriptor-based and learning-based methods on the FIRE dataset. Specifically, GPO-DCN achieves the lowest TRE (\textbf{2.352}\,px) and consistently ranks highest across all AUC thresholds(0.938@25\,px and 0.964@50\,px). Compared to the best-performing descriptor-based method (GeoFormer~\cite{liu2023geoformer}, TRE~6.201\,px) and the strongest learning-based method (RetinaRegNet~\cite{sivaraman2025retinaregnet}, TRE~2.766\,px), GPO-DCN reduces alignment error by 3.8\,px and 0.4\,px, respectively. These gains reflect the benefit of anatomically guided nodes and KNN-based blending in modelling local deformations beyond homography or uniform grids. Moreover, GPO-DCN attains higher AUC values at all thresholds (0.906@15\,px vs.\ 0.625 for GeoFormer), indicating improved spatial precision and robustness to outlier errors through anatomically guided node placement.

By contrast, GPO-GCN achieves slightly higher TRE (2.649\,px) and lower AUCs (0.914@25\,px), but still exceeds other baselines, underscoring the benefit of the iterative Gaussian-primitive optimization. However, its uniform-grid sampling may miss vessel junctions or small-scale variations, resulting in lower AUC compared to GPO-DCN—especially at moderate thresholds  (e.g., AUC@25\,px = 0.914). Consequently, GPO-DCN’s descriptor-based node acquisition preserves more fine-grained vascular structure and ensures more robust alignment.

\begin{table}[t!]
\caption{Quantitative comparison on the FIRE dataset. Methods are categorized into Descriptor-based methods (left) and Learning \& Iterative Optimization methods (right). Best results are in \textbf{bold}, second-best are \underline{underscored}.}
\label{tab:comparison}
\centering
\resizebox{\textwidth}{!}{%
\begin{tabular}{l c ccc @{\hskip 1em} l c ccc}
    \toprule
    \multicolumn{5}{c}{\textit{Descriptor-based Methods}} & \multicolumn{5}{c}{\textit{Learning \& Iterative Optimization Methods}} \\
    \cmidrule(lr){1-5} \cmidrule(lr){6-10}
    \multirow{2}{*}[-0.5ex]{\textbf{Method}} & \multirow{2}{*}[-0.5ex]{\textbf{TRE} $\downarrow$} & \multicolumn{3}{c}{\textbf{AUC} $\uparrow$} & \multirow{2}{*}[-0.5ex]{\textbf{Method}} & \multirow{2}{*}[-0.5ex]{\textbf{TRE} $\downarrow$} & \multicolumn{3}{c}{\textbf{AUC} $\uparrow$} \\
    \cmidrule(lr){3-5} \cmidrule(lr){8-10}
    & & @15 & @25 & @50 & & & @15 & @25 & @50 \\
    \midrule
    XFeat~\cite{potje2024xfeat} & 10.858 & 0.560 & 0.637 & 0.794 & GraDIRN~\cite{qiu2022GraDIRN} & 6.344 & 0.657 & 0.774 & 0.885 \\
    R2D2~\cite{revaud2019r2d2} & 7.926  & 0.553 & 0.701 & 0.850 & PDD-Net~\cite{heinrich2019pdd} & 5.765 & 0.688 & 0.792 & 0.893 \\
    LightGlue~\cite{lindenberger2023lightglue} & 7.802  & 0.575 & 0.710 & 0.855 & VoxelMorph++~\cite{heinrich2022voxelmorph++} & 5.400 & 0.710 & 0.808 & 0.902 \\
    SuperPoint~\cite{detone2018superpoint} & 6.641  & 0.612 & 0.757 & 0.879 & VR-Net~\cite{jia2021vrnet} & 4.974 & 0.705 & 0.823 & 0.911 \\
    Glampoints~\cite{truong2019glampoints} & 6.608  & 0.595 & 0.757 & 0.879 & RetinaRegNet~\cite{sivaraman2025retinaregnet} & 2.766 & 0.852 & 0.910 & \underline{0.955} \\
    SuperRetina~\cite{liu2022superretina} & 6.382  & 0.622 & 0.767 & 0.884 & \multicolumn{5}{l}{\rule[0.5ex]{0.55\linewidth}{0.4pt}} \\
    RoMa~\cite{edstedt2024roma} & 6.388  & 0.605 & 0.763 & 0.881 & GPO-GCN (Ours) & \underline{2.649} & \underline{0.888} & \underline{0.914} & 0.926 \\
    GeoFormer~\cite{liu2023geoformer} & 6.201  & 0.625 & 0.770 & 0.887 & GPO-DCN (Ours) & \textbf{2.352} & \textbf{0.906} & \textbf{0.938} & \textbf{0.964} \\
    \bottomrule
\end{tabular}
}
\end{table}

\subsubsection{Qualitative Analysis}
\begin{figure}[t]
    \centering
    \includegraphics[width=\textwidth]{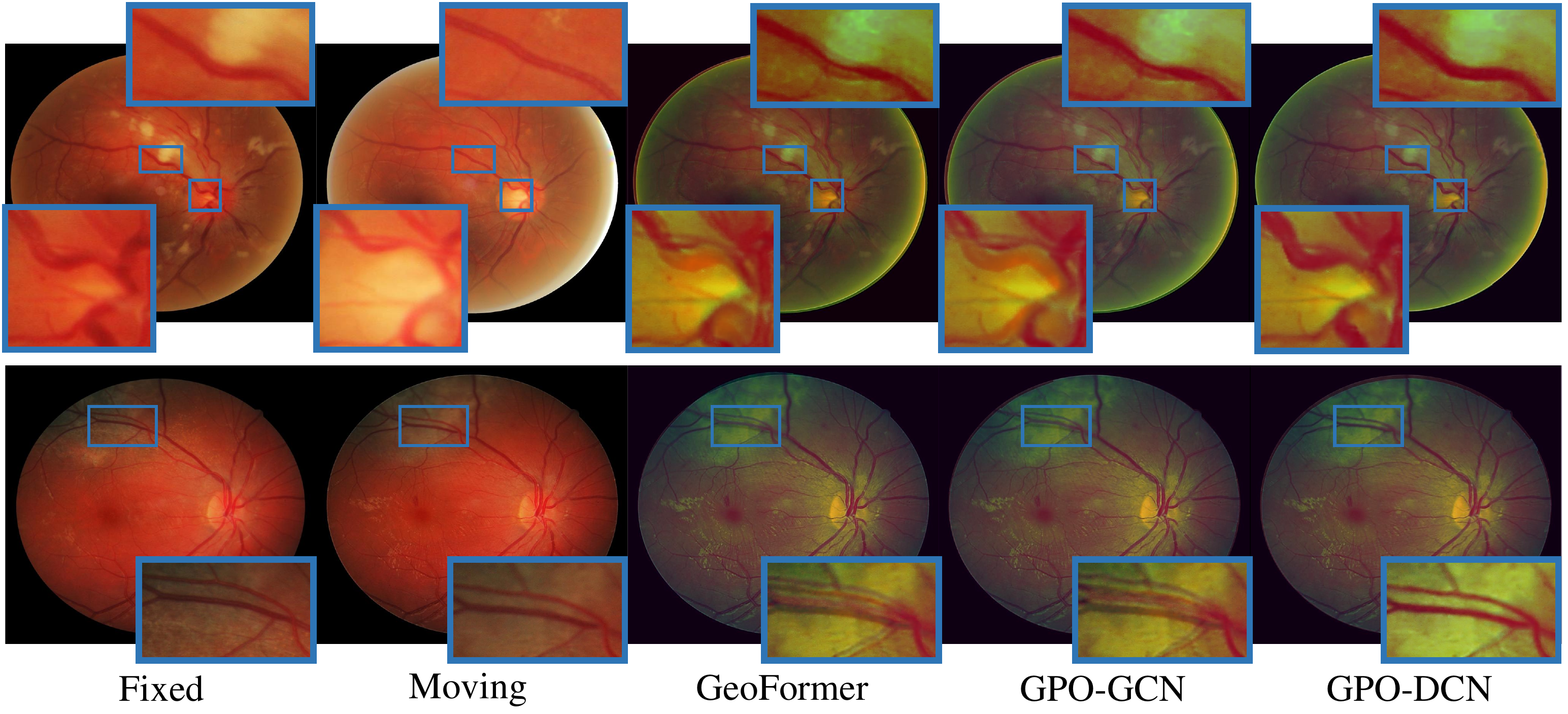}
    \caption{Qualitative comparison on the FIRE dataset.}
    \label{fig:qualitive}
\end{figure}
Fig~\ref{fig:qualitive} compares \( I_m \), \( I_f \), and registration outputs from GeoFormer, GPO-GCN, and GPO-DCN on two challenging retinal cases. In \emph{top row}, we highlight a region at the optic disc where vessel geometry is highly tortuous and subject to complex localized deformations. GeoFormer and GPO-GCN struggle to preserve these fine details, leading to partial misalignment. In contrast, GPO-DCN leverages descriptor-based control nodes for the message passing framework to capture complex localized distortions more accurately, preserving vessel continuity. The \emph{bottom row} highlights shadowing artifacts that obscure sections of the vasculature. Here, both GeoFormer and GPO-GCN exhibit residual misalignment in shadowed regions, whereas GPO-DCN demonstrates tighter correspondence of vascular edges. By iteratively refining Gaussian primitives near salient features, GPO-DCN effectively compensates for local intensity variations, resulting in sharper vessel alignment even under low contrast.

\subsubsection{Ablation Studies}
\label{sec:ablation}
\begin{figure}[t]
    \centering
    \begin{minipage}{0.33\textwidth}
        \includegraphics[width=\linewidth]{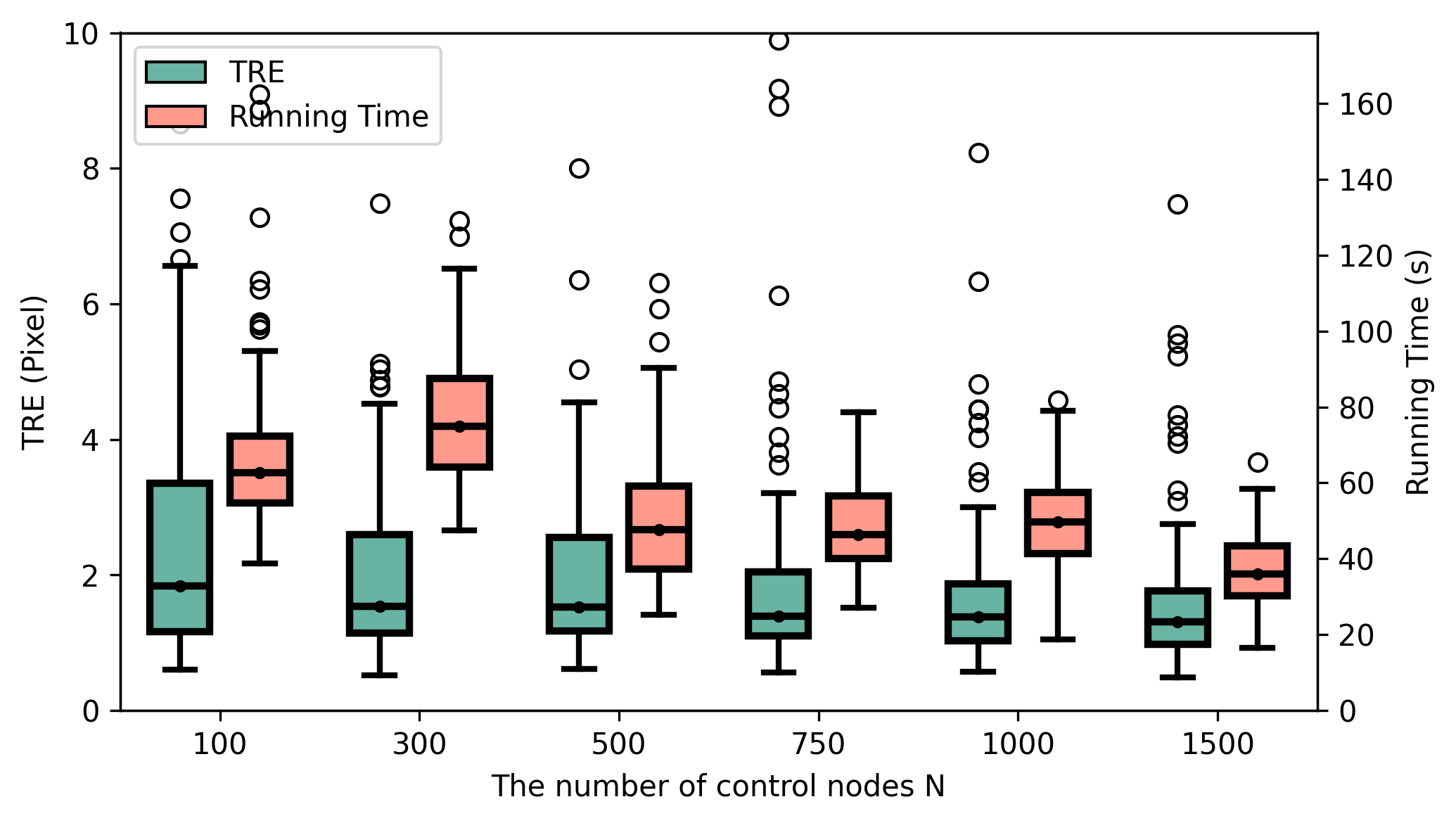}
    \end{minipage}
    \hfill
    \begin{minipage}{0.32\textwidth}
        \includegraphics[width=\linewidth]{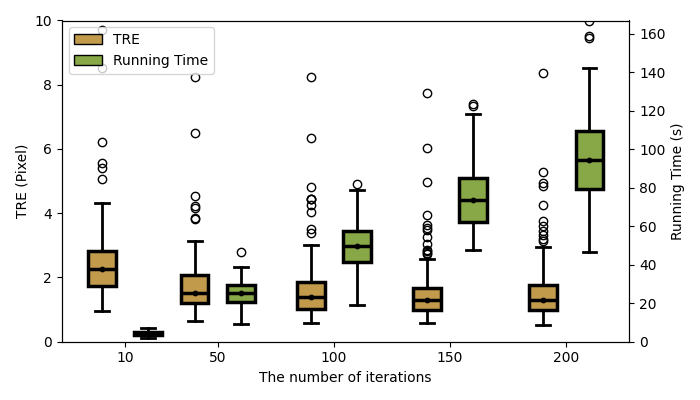}
    \end{minipage}
    \hfill
    \begin{minipage}{0.33\textwidth}
        \includegraphics[width=\linewidth]{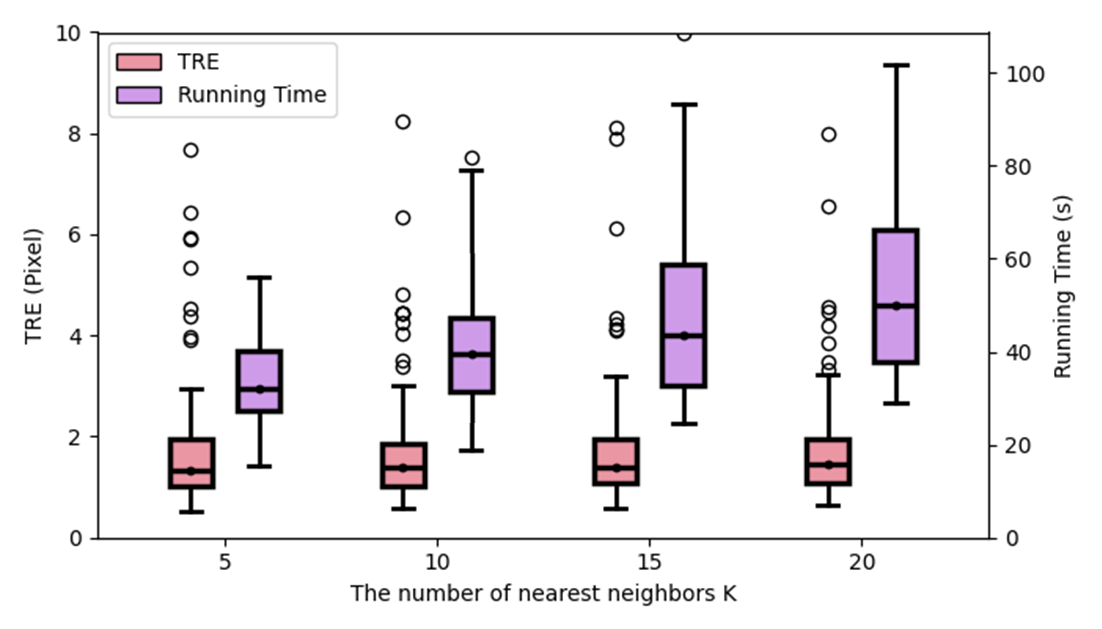}
    \end{minipage}
    \caption{Ablation study on key parameters on TRE and running time. \textbf{Left:} Influence of the number of control nodes \(N\). \textbf{Middle:} Effect of the number of iterations. \textbf{Right:} Impact of the number of nearest neighbors \(K\).}
    \label{fig:ablation}
\end{figure}

We conducted a three-way ablation to examine how the number of control nodes \(N\), the number of nearest neighbors \(K\), and the number of optimization iterations \(\tau\) affect both accuracy and runtime (Fig.\,\ref{fig:ablation}).

\textbf{Number of Control Nodes \(\mathbf{N}\):}
Increasing \(N\) from 300 to 1000 reduces the median TRE from \(\sim2.60\)\,px to \(\sim2.35\)–2.40\,px but raises runtime from \(\sim18\)s to \(\sim34\)s. Beyond 1000 nodes, further improvements (\(\sim2.22\)–2.35\,px) come at the expense of a longer runtime (\(\sim45\)s), yielding diminishing returns.

\textbf{Number of Nearest Neighbors \(\mathbf{K}\):}
For \(K=5\), the average TRE remains at \(\sim2.60\)–2.65\,px. Increasing to \(K=10\) lowers it to \(\sim2.40\)–2.45\,px, with a moderate \(\sim30\)s runtime. Although going to \(K=15\) or \(K=20\) can yield minor accuracy gains (\(\sim2.35\)–2.38\,px), runtime increases by 20–30\%. Thus, \(K=10\) strikes the best balance between precision and efficiency.

\textbf{Number of Iterations \(\boldsymbol{\tau}\):}
With \(\tau=50\), the median TRE hovers around 2.55–2.60\,px in \(\sim15\)s. Doubling to \(\tau=100\) reduces TRE to \(\sim2.40\)–2.45\,px (a 0.1–0.2\,px gain) at \(\sim30\)s. Beyond 100 iterations, further gains (\(\le0.05\)–0.07\,px) come at a steep runtime cost (\(\sim45\)s or more).

\noindent Overall, \(N=1000\), \(K=10\), and \(\tau=100\) achieve a median TRE of \(\sim2.3\)–2.4\,px within \(\sim30\)s, offering a favorable trade-off between alignment accuracy and computational overhead.

\section{Conclusion}
 \label{sec:conclusion} We introduced GPO, an iterative deformable registration framework that addresses the critical challenge of gradient signal dilution in retinal images. By leveraging descriptor-based control nodes (or uniform-grid nodes) to blend and propagate their local transformations via KNN-weighted Gaussian primitives, GPO mitigates vanishing gradients in homogeneous regions while accurately modelling localized vessel deformations. On the FIRE dataset, GPO outperforms conventional homography-based and learning-based methods in both alignment accuracy and robustness. Future work will explore extending GPO to multimodal retinal registration and incorporating multi-scale optimization strategies for broader medical imaging applications.

\begin{credits}
\subsubsection{\discintname}
The authors have no competing interests to declare that are relevant to the content of this article.
\end{credits}

\bibliographystyle{splncs04}
\bibliography{Paper-3875.bib}

\end{document}